\documentclass[11pt,a4paper]{article}
\usepackage[hyperref]{acl2018}
\usepackage{amssymb}
\usepackage{times}
\usepackage{latexsym}
\usepackage{url}
\usepackage{amsmath}
\usepackage{booktabs}
\usepackage{multirow}
\usepackage{enumerate}
\usepackage{mathtools}
\DeclareMathAlphabet{\pazocal}{OMS}{zplm}{m}{n}
\usepackage{rotating}

\usepackage{tikz}
\usetikzlibrary{positioning,fit}
\usetikzlibrary{overlay-beamer-styles}

\usetikzlibrary{arrows,shapes}
\usepackage{rotating}

\usepackage{comment}
\usepackage{flowchart}
\usetikzlibrary{arrows}

\aclfinalcopy % Uncomment this line for the final submission
\def\aclpaperid{844} %  Enter the acl Paper ID here

\usepackage{xspace}

\DeclareMathOperator \real{\mathbb{R}}

\newcommand{\Ls}{\mathcal{L}}

\newcommand{\Us}{\pazocal{U}}

\newcommand{\red}[1]{\textcolor{red}{#1}}
\newcommand{\blue}[1]{\textcolor{blue}{#1}}

\newcommand{\Ni}{({\em i})~}
\newcommand{\Nii}{({\em ii})~}

\newcommand{\ie}{{\em i.e.,}\xspace}
\newcommand{\eg}{{\em e.g.,}\xspace}

\newenvironment{noindlist}
 {\begin{list}{\labelitemi}{\leftmargin=0em \itemindent=1em}}
 {\end{list}}

\newcommand{\gridcnn}{\nobreak{Neural Grid}}

\newcommand{\gridcnnlex}{\nobreak{Lex. Neural Grid}}

\newcommand{\wsj}{\textsc{wsj}}
\newcommand{\gridallnoun}{\nobreak{Grid}}

\title{Coherence Modeling of Asynchronous Conversations: \\A Neural Entity Grid Approach}

\author{Tasnim Mohiuddin\thanks{All authors contributed equally.} \and Shafiq Joty$^*$ \\
	Nanyang Technological University\\  
  {\tt \{mohi0004,srjoty\}@ntu.edu.sg} \\\And
  Dat Tien Nguyen$^*$ \\
  University of Amsterdam \\
  {\tt t.d.nguyen@uva.nl} \\}

\date{}

\begin{document}
\maketitle

\begin{abstract}

We propose a novel coherence model for written asynchronous conversations (e.g., forums, emails), and show its applications in coherence assessment and thread reconstruction tasks. We conduct our research in two steps. First, we propose improvements to the recently proposed neural entity grid model by lexicalizing its entity transitions. Then, we extend the model to asynchronous conversations by incorporating the underlying conversational structure in the entity grid representation and feature computation. Our model achieves state of the art results on standard coherence assessment tasks in monologue and conversations outperforming existing models. 
We also demonstrate its effectiveness in reconstructing thread structures.  

% and 
%A further evaluation demonstrates the utility of our neural coherence model for the task of thread reconstruction in forum conversation.

\end{abstract}

\section{Introduction}

Sentences in a text or a conversation do not occur independently, rather they are connected to form a coherent discourse that is easy to comprehend. \textbf{Coherence models} are computational models that can distinguish a coherent discourse from incoherent ones. It has ranges of applications in text generation, summarization, and coherence scoring. 

%that can distinguish a coherent from incoherent texts (resp. conversations) 
%What distinguishes a well-written text from a random sequence of sentences is that it binds the sentences together to express a meaning as a whole. Together the sentences form a coherent discourse (a monologic or a conversational) that is easy to comprehend. 

Inspired by formal theories of discourse, a number of coherence models have been proposed  \cite{Barzilay:2008,Lin:2011,li-jurafsky:2017}. The \textbf{entity grid} model \cite{Barzilay:2008} is one of the most popular coherence models that has received much attention over the years. As exemplified in {Table} \ref{table:doc}, the model represents a text by a grid that captures how grammatical roles of different discourse entities (\eg\ nouns) change from one sentence to another in the text. The grid is then converted into a feature vector containing probabilities of local entity transitions, enabling machine learning models to measure the degree of coherence. Earlier extensions of this basic model incorporate entity-specific features \cite{Elsner:2011}, multiple ranks \cite{Feng:2012}, and coherence relations \cite{Feng:2014}.

Recently, \citet{dat-joty:2017} proposed {a} neural {version} of the grid models. Their model first transforms the grammatical roles in a grid into their distributed representations, and  employs a convolution operation over it to model entity transitions in the distributed space. The spatially  max-pooled  features from the convoluted features are used for coherence scoring. This model achieves state-of-the-art results in standard evaluation tasks on the Wall Street Journal (\wsj) corpus.

\begin{figure}[t!]
{\small
\begin{enumerate}\setlength\itemsep{-0.2em}
 \item[s$_0$:] \textbf{LDI} Corp., Cleveland, said it will offer \$50 million in commercial {\bf paper} backed by lease\-rental receivables.  
 \item[s$_1$:] The program matches funds raised from the sale of the commercial {\bf paper} with small to medium-sized leases.
 \item[s$_2$:] \textbf{LDI} termed the {\bf paper} ``non-recourse financing'', meaning that investors  would be repaid from the lease receivables, rather than directly by LDI Corp.
 \item[s$_3$:]\textbf{LDI} leases and sells data-processing, telecommunications and other high-tech equipment.
\end{enumerate}}
\vspace{-1em}
\end{figure}
%\vspace{-0.3em}
\begin{table}[t!]
\centering
\resizebox{0.99\linewidth}{!}{
{\small
\begin{tabular}{r|ccccccccccccccccc}
& {\rotatebox{90}{INVESTORS}} & {\rotatebox{90}{MILLION}} & {\rotatebox{90}{FUNDS}} & {\rotatebox{90}{EQUIPMENT}} & {\rotatebox{90}{CORP.}} & {\rotatebox{90}{\textbf{PAPER}}} & {\rotatebox{90}{SALE}} & {\rotatebox{90}{TELECOMM.}} & {\rotatebox{90}{LEASE}}& {\rotatebox{90}{PROGRAM}} & {\rotatebox{90}{CLEVELAND}} & {\rotatebox{90}{RECEIVABLES}} & {\rotatebox{90}{LEASES}} & {\rotatebox{90}{DATA-PROCESS.}} & {\rotatebox{90}{\textbf{LDI}}} &  {\rotatebox{90}{NON-RECOURSE}}  \\
\toprule
s$_0$ & $-$ &O & $-$ & $-$ & S & X &  $-$ & $-$ & $-$ & $-$ &X &X & $-$& $-$ & X & $-$ \\
s$_1$ & $-$&$-$&O&$-$&$-$&X&X&$-$&$-$&S&$-$&$-$&X&$-$&$-$&$-$\\
s$_2$ & S&$-$&$-$&$-$&X&S&$-$&$-$&X&$-$&$-$&X&$-$&$-$&S&X\\
s$_3$ & $-$&$-$&$-$&O&$-$&$-$&$-$&X&$-$&$-$&$-$&$-$&$-$&X&S&$-$\\
\bottomrule
\end{tabular}
}
}
\vspace{-0.3em}
\caption{Entity grid representation (bottom) for a document (top) from the \wsj\ corpus.} %(doc id: 1352). Grid cells correspond to grammatical roles: subjects (S), objects (O), other (X), and absent (--).}
\label{table:doc}
\vspace{-0.3em}
\end{table}

Although the neural grid model effectively captures long entity transitions, it is still limited in  that it does not consider any lexical information regarding the entities, thereby, fails to distinguish between entity types. Although the extended neural grid considers entity features like named entity and proper mention, it requires an explicit feature extraction step, which can prevent us to transfer the model to a  resource-poor language or domain. %Second, the current model employs only one convolution operation at a time, thus cannot extract and combine features from varying window sizes.   

Apart from these limitations, previous research on coherence models has mainly focused on monologic discourse (\eg\ news article). The only exception is the work of \citet{Elsner:2011-chat}, who applied coherence models to the task of conversation disentanglement in \textbf{synchronous} conversations like phone and chat conversations.

With the emergence of Internet technologies, \textbf{asynchronous} communication media like {emails}, {blogs,} and {forums} have become a commonplace for discussing events and issues, seeking answers, and sharing personal experiences. {Participants in these media interact with each other asynchronously, by writing at different times}. %KA: this line may need rewriting  %Effective processing of these conversational texts can lead to many useful applications \cite{Carenini11book}. 
We believe coherence models for asynchronous conversations can help many downstream applications in these domains. For example, we will demonstrate later that coherence models can be used to predict the underlying thread structure of a conversation, which provides crucial information for building effective conversation summarization systems \cite{Carenini08} and community question answering systems \cite{cedeno-et-al-acl-15}.

%like conversation summarization and  community question answering (cQA) system 

%can be of great value for both organizations and individuals  For example, conversation summarization systems can help providing an overview of what people are saying and what opinions are being expressed. Likewise, a community question answering (cQA) system can provide direct answers to questions by analyzing thousands of comments posted in community forum sites. 

To the best of our knowledge, none has studied the problem of coherence modeling in asynchronous conversation before. Because of its asynchronous nature, information flow in these conversations is often not sequential as in monologue or synchronous conversation. This poses a novel set of challenges for discourse analysis models \cite{Joty:2013,louis2015conversation}. For example, consider the forum conversation in Figure \ref{fig:thread-tree}(a). It is not obvious how a coherence model like the entity grid can represent the conversation, and use it in downstream tasks effectively.

%As a result, discourse structures such as topic structure, coherence structure, and conversational structure in these conversations exhibit different properties than what we observe in monologue or synchronous conversation \cite{Joty:2013,louis2015conversation}. 

%For instance, entity grids constructed from the temporal order of the utterances may not provide the best representation for the  conversation.

In this paper we aim to remedy the above limitations of existing models in two steps. First, we propose improvements to the existing neural grid model by \emph{lexicalizing} its entity transitions. We propose methods based on word embeddings to achieve better generalization with the lexicalized model. %and \Nii we perform multiple convolution-pooling operations concurrently with different window sizes to give the model more expressive power. 
Second, we adapt the model to asynchronous conversations by incorporating the underlying \emph{conversational structure} in the grid representation and subsequently in  feature computation. For this, we propose a novel grid representation for asynchronous conversations, and adapt the convolution layer of the neural model accordingly.

%s: one at the path-level and the other at the tree-level.

{We evaluate our approach on two discrimination tasks. The first task is the standard one, where we assess the models based on their performance in discriminating an original document from its random permutation. In our second task, we ask the models to distinguish an original document from its inverse order of the sentences.} 
For our adapted model to asynchronous conversation, we also evaluate it on \emph{thread reconstruction}, a task specific to asynchronous conversation. We performed a series of experiments, and our main findings are: 

%\vspace{-0.2em}
\begin{enumerate}[(a)] \setlength\itemsep{-0.2em}

\item {Our experiments on the \wsj\ corpus validate the utility of our proposed extension to the existing neural grid model, yielding absolute $F_1$ improvements of up to 4.2\% in the standard task and up to 5.2\% in the inverse-order discrimination task, setting a new state-of-the-art.}  

\item {Our experiments on a forum dataset show that our adapted model that considers the conversational structure outperforms the temporal baseline by more than 4\% $F_1$  in the standard task and by about 10\% $F_1$  in the inverse order discrimination task.}

{\item When applied to the thread reconstruction task, our model achieves promising results outperforming several strong baselines.}   

%\vspace{-0.2em}
\end{enumerate}

{We have released our source code and datasets at \url{https://ntunlpsg.github.io/project/coherence/n-coh-acl18/}} 

%\blue{In the rest of the paper, after discussing related work in Section \ref{sec:rel-work}, we present our lexicalized entity grid model in Section \ref{sec:neural-grids}. In Section \ref{sec:coh-model-conv}, we present our proposed coherence model for asynchronous conversations. After discussing results on monologue for our lexicalized grid model in Section \ref{sec:exp}, we present our results on forum conversations in Section \ref{sec:exp2}. Finally, we summarize our contributions with future directions in Section \ref{sec:con}.}  

%The experiments and analysis of results are presented in Section \ref{sec:exp}. 

%used in our experiments for research purposes.%\footnote{Available from \url{https://github.com/datienguyen/cnn_coherence/}}     

%on the standard evaluation tasks (\ie\ discrimination and insertion).

%, generating a type of conversational discourse where information flow is often not sequential as in monologue or synchronous conversation. 

%show that our neural models consistently improve over their non-neural counterparts (\ie\ existing entity grid models) yielding absolute gains of about $4\%$  on discrimination, up to $2.5\%$ on insertion, and more than $4\%$ on summary coherence rating. %Furthermore, our model achieves state of the art results in all these tasks.

\section{Background}
\label{sec:rel-work}
%\vspace{-0.3em}
In this section we give an overview of existing coherence models. In the interest of coherence, we defer description of the neural grid model \cite{dat-joty:2017} until next section, where we present our extension to this model. 

%\vspace{-0.2em}
\subsection{Traditional Entity Grid Models} \label{egrids}

Introduced by \citet{Barzilay:2008}, the \textbf{entity grid} model represents a text by a two-dimensional matrix. %that captures transitions of discourse entities across sentences. 
As shown in Table \ref{table:doc}, the rows  correspond to sentences, and the columns correspond to entities (noun phrases). Each entry $E_{i,j}$ represents the syntactic role that entity $e_j$ plays in sentence $s_i$, which can be one of: subject (S), object (O), other (X), or absent ({--}). {In cases where an entity appears more than once with different grammatical roles in the same sentence, the role with the highest rank (S $\succ$ O $\succ$ X) is considered.}

Motivated by the Centering Theory \cite{Grosz:1995}, the model considers \textbf{local entity transitions} as the deciding patterns for assessing coherence. A {local entity transition} of length $k$ is a sequence of $\{$S,O,X,--$\}^k$, representing grammatical roles played by an entity in $k$ consecutive sentences. Each grid is represented by a vector of $4^k$ transition probabilities computed from the grid. %For example, the probability of the transition $[X--]$ in Figure~\ref{table:grid} is $0.079$, computed as a ratio of its frequency (i.e., 3) divided by the total number of transitions of length three (i.e., 38) in the entity grid. 
To distinguish between transitions of important entities from  unimportant ones, the model   considers the \emph{salience} of the entities, which is measured by their occurrence frequency in the document. With the feature vector representation, coherence assessment task is formulated as a ranking problem in a SVM preference ranking framework \cite{Joachims:2002}. \citet{Barzilay:2008} showed significant improvements in two out of three evaluation tasks when a coreference resolver is used to identify coreferent entities in a text.

\citet{Elsner:2011} show improvements to the grid model by including non-head nouns as entities. %with role `other (X)' in the grid. 
Instead of employing a coreference resolver, they match the nouns to detect coreferent entities.
%\citet{Elsner:2011} proposed several improvements to the grid model. They first show improvement by including non-head nouns (\ie\ nouns that do not head NPs) as entities with role `other (X)' in the grid. Instead of employing a coreference resolver, they match the nouns to detect coreferent entities. This provides more information to the standard grid, while also maximizing coreference recall \cite{Elsner:2010:SHC}. 
They demonstrate further improvements by extending the grid to distinguish between entities of different types. They do so by incorporating entity-specific features like named entity, noun class and modifiers. %\footnote{Recall that the original entity grid uses only one entity-specific feature, which is \emph{Salience}.} %These extensions led to the best results reported so far.
%\citet{Feng:2012} improved the ranking scheme of the basic entity grid model by learning not only from original document and its permutations but also from preferences among the permutations. 
\citet{Lin:2011} model transitions of discourse roles for entities as opposed to their grammatical roles. They instantiate discourse roles by discourse relations in Penn Discourse Treebank \cite{PRASAD08.754}. In a follow up work, \citet{Feng:2014} trained the same model but using relations derived from deep discourse structures annotated with Rhetorical Structure Theory \cite{mann1988rhetorical}. 
 
% They consider noun phrases (NP) as {entities}, and employ a coreference resolver to detect mentions of the same entity (\eg\ \emph{Obama} and \emph{the president}).  Consider for example the column  for the entity \emph{Eaton} in Figure \ref{table:doc}, [\textsc{X - - S}]. It indicates that Eaton plays an `other' and a `subject' role in the first and the last sentences, respectively, and it is absent from the rest of the sentences. 

%\subsection{Neural Entity Grid Models} \label{egrids}

%In the simplest scenario, $\Vs$ contains $\{S,O,X,-\}$. However, we will see in Section \ref{sec:ext-grid-cnn} that as we include more entity-specific features, $\Vs$ can contain more symbols. We consider $E$ a model parameter to be learned by backpropagation on a given task. We can initialize $E$ randomly or using pretrained vectors trained on a general coherence task.

%\vspace{-0.1em}
\subsection{Other Existing Models} \label{other-models}

\citet{Guinaudeau:2013} proposed a \textbf{graph-based} unsupervised method. %for assessing coherence. 
They convert an entity grid into a bipartite graph consisting of two sets of nodes, representing sentences and entities, respectively. {The edges are assigned weights based on the grammatical role of the entities in the respective sentences. They perform one-mode projections to transform the bipartite graph to a directed graph containing only sentence nodes. The coherence score of the document is then computed as the average \emph{out-degree} of sentence nodes.} 

\citet{Louis:2012:CMB} introduced a coherence model based on \textbf{syntactic patterns} by assuming that sentences in a coherent text exhibit certain syntactic regularities. They propose a local coherence model that captures the co-occurrence of structural features in
adjacent sentences, and a global model based on a hidden Markov model, which learns the global syntactic patterns from clusters of sentences with similar syntax.

%The  weights reflects the grammatical role
%between a sentence node and an entity node reflects the grammatical role of the entity in the sentence. %; they use  weights of $3$ for S, $2$ for O, $1$ for X, and $0$ for --. 
%To model entity transitions, they perform one-mode projections to transform the bipartite graph to a directed graph containing only sentence nodes -- two sentences are connected if they share at least one entity in common. The coherence score of the document is then computed as the average \emph{out-degree} of sentence nodes. 

%More recent approaches use neural networks. 

\citet{li-hovy:EMNLP20142} proposed a \textbf{neural} framework to compute the coherence score of a document by estimating coherence probability for every window of three sentences. They encode each sentence in the window using either a recurrent or a recursive neural network. To get a document-level coherence score, they sum up the window-level log probabilities. \citet{li-jurafsky:2017} proposed two encoder-decoder models augmented with latent variables for both  coherence evaluation and discourse generation. Their first model incorporates global discourse information (topics)  by feeding the output of a sentence-level HMM-LDA model \cite{pmlr-v2-gruber07a} into the encoder-decoder model. Their second model is trained end-to-end with variational inference. %\cite{KingmaW13}.

{In our work, we take an entity-based approach, and extend the neural grid model  proposed recently by \citet{dat-joty:2017}}.

%, similar to the method used for learning variational auto encoders \cite{KingmaW13}.  

%In their experiments, they use three different weighting schemes for edges: \Ni {\it binary}, whether the sentences have any entity in common, \Nii {\it weighted}, where edge weight is the number of common entities, and \Niii {\it syntactic}, where edge weights are the sum of the weights (reflecting grammatical roles of common entities) in the bipartite graph.  

\section{Extending Neural Entity Grid} 
\label{sec:neural-grids}

In this section we first briefly describe the neural entity grid model proposed by  \citet{dat-joty:2017}. {Then,} we propose our extension to this model that leads to improved performance. {We present our coherence model for asynchronous conversation in the next section.}

\subsection{Neural Entity Grid}

% neuralize existing grid models.  
%\citet{dat-joty:2017} recently proposed \textbf{neural entity grid} models to measure text coherence. 

\begin{figure}[t]
\centering
\includegraphics[width=0.48\textwidth]{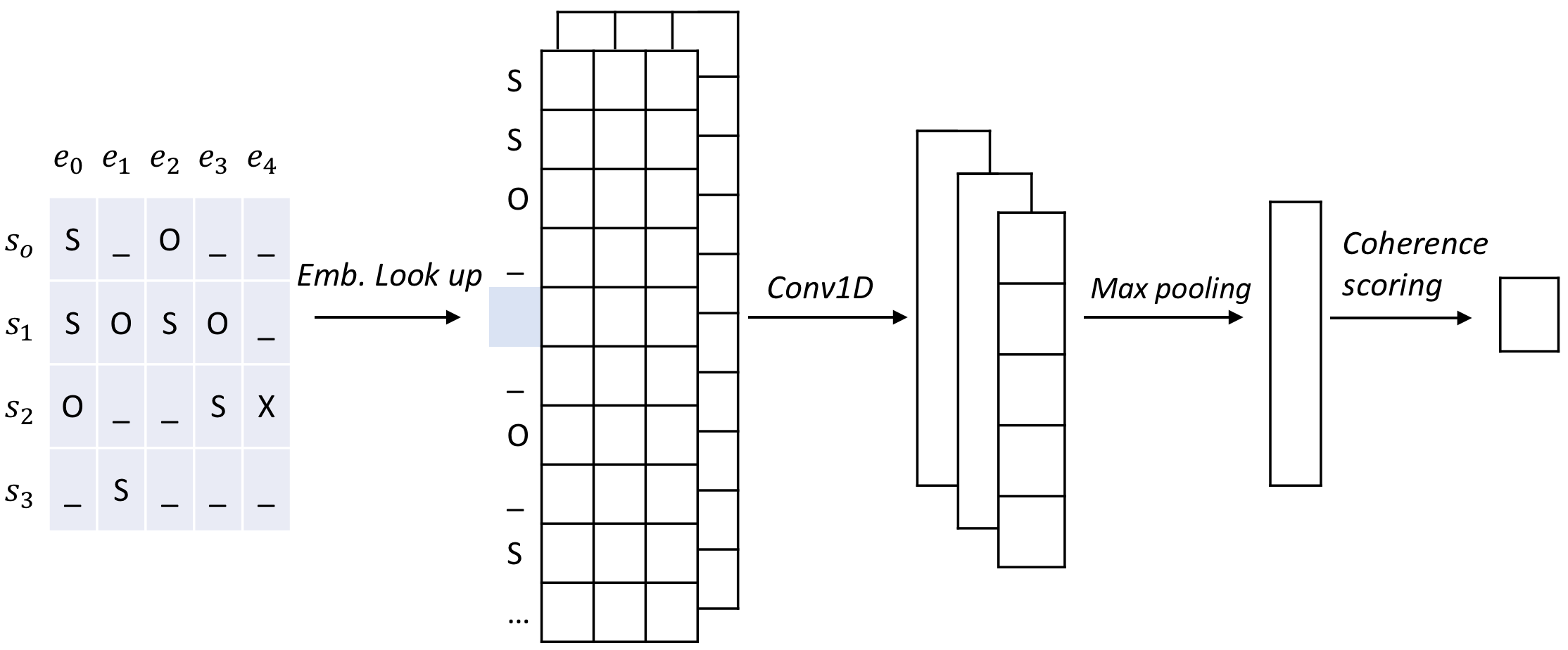}
\vspace{-1em}
\caption{Neural entity grid model proposed by \citet{dat-joty:2017}. The model is trained using a pairwise ranking approach with shared parameters for positive and negative documents.}
\label{fig:neural-grid}
%\vspace{-0.1em}
\end{figure}

{Figure \ref{fig:neural-grid} depicts the neural grid model of \citet{dat-joty:2017}}. Given an entity grid $E$, they first transform each entry $E_{i,j}$ (a grammatical role) into a distributed representation of $d$ dimensions  by looking up a shared embedding matrix $M$ $\in$ $\real^{|G| \times d}$, where $G$ is the vocabulary of possible grammatical roles, \ie\ $G=\{S,O,X,-\}$. Formally, the look-up operation can be expressed as:    

\vspace{-1.0em}
\begin{equation}
 L = \Big[ M({E_{1,1}}) \cdots M({E_{i,j}}) \cdots M({E_{I,J}})  \Big] \label{lookup}
\end{equation}
%\vspace{-1.3em}

\noindent where $M({E_{i,j}})$ refers to the row in $M$ that corresponds to grammatical role $E_{i,j}$, and $I$ and $J$ are the number of rows (sentences) and columns (entities) in the entity grid, respectively. The result of the look-up operation is a tensor $L \in \real^{I \times J \times d}$, which is fed to a convolution layer to model local entity transitions in the distributed space. 

The convolution layer of the neural network composes patches of entity transitions into high-level abstract features by treating entities  independently (\ie\ 1D convolution). Formally, it applies a \emph{filter} $\mathbf{w} \in \real^{m.d}$ to each local entity transition of length $m$ to generate a new abstract feature $z_i$:  

%\vspace{-0.8em}
\begin{equation}
z_i = h(\mathbf{w}^T L_{i:i+m, j} + b_i) \label{eq:conv}
\end{equation}  
%\vspace{-1.2em}

\noindent where $L_{i:i+m, j}$ denotes concatenation of $m$ vectors in $L$ for entity $e_j$, $b_i$ is a bias term, and $h$ is a nonlinear activation function. Repeated application of this filter to each possible $m$-length transitions of different entities in the grid generates a \emph{feature map}, $\mathbf{z}^i = [z_1, \cdots, z_{I.J+m-1}]$. This process is repeated $N$ times with $N$ different filters to get $N$ different feature maps, $[\mathbf{z}^1, \cdots, \mathbf{z}^N]$. A \emph{max-pooling} operation is then applied to extract the most salient features from each feature map:

%\vspace{-0.8em}
\begin{equation}
\mathbf{p} = [\mu_l(\mathbf{z}^1), \cdots, \mu_l(\mathbf{z}^N)] \label{max_pool}
\end{equation}
%\vspace{-1.1em}

\noindent where $\mu_l(\mathbf{z}^i)$ refers to the $\max$ operation applied to each non-overlapping window of $l$ features in the feature map $\mathbf{z}^i$. Finally, the pooled features are used  in a linear layer to produce a  \emph{coherence score}: 

%\vspace{-0.8em}
\begin{equation}
y = \mathbf{u}^T \mathbf{p} + b \label{dense} 
\end{equation}
%\vspace{-1.2em}

\noindent where $\mathbf{u}$ is the weight vector and ${b}$ is a bias term. The model is trained with a \emph{pairwise ranking} loss based on ordered training pairs $(E_i, E_j)$:  

\vspace{-1.0em}
\begin{equation}
\Ls(\theta)=  \max \{0, 1 - \phi(E_i|\theta) + \phi(E_j|\theta)\} \label{loss}
\end{equation}
%\vspace{-1.4em}

\noindent where entity grid $E_i$ exhibits a higher degree of coherence than grid $E_j$, and $y=\phi(E_k|\theta)$ denotes the transformation  of  input grid $E_k$ to a coherence score $y$ done by the model with parameters $\theta$. {We will see later that such ordering of documents (grids) can be obtained automatically by permuting the original document. Notice that the network shares its parameters ($\theta$) between the positive ($E_i$) and the negative ($E_j$) instances in a pair.} 

%to obtain $\phi(G_i|\theta)$ and $\phi(G_j|\theta)$ from a pair of input grids $(G_i, G_j)$.}

Since entity transitions in the convolution step are modeled in a continuous space, it can effectively capture longer transitions compared to traditional grid models. Unlike traditional grid models that compute transition probabilities from a \emph{single} grid, convolution filters and role embeddings in the neural model are learned from all training instances, which helps the model to generalize well.

{Since the abstract features in the feature maps are generated by convolving over role transitions of different entities in a document, the model implicitly considers relations between entities in a document, whereas transition probabilities in traditional entity grid models are computed without considering any such relation between entities. Convolution over the entire  grid also incorporates \emph{global} information (\eg\ topic) of a discourse.} %\citet{dat-joty:2017} show that their model achieves state-of-the-art results in the standard evaluation tasks on the \wsj\ dataset.}

%
%demonstrate empirically that the neural grid model achieves state-of-the-art results in the standard evaluation tasks on the \wsj\ benchmark dataset. 

%\blue{Since the abstract features in the feature maps are generated by convolving over role transitions of different entities in a document, the model implicitly considers relations between entities in a document, whereas transition probabilities in traditional entity grid models are computed without considering any such relation between entities. Convolution over the entire entity grid also allows the model to incorporate global information (\eg\ topic) of a discourse.}   

%\red{revised up to this}

%\subsection{Limitations}
\subsection{Lexicalized Neural Entity Grid}

Despite its effectiveness, the neural grid model presented above has a limitation.
%number of limitations. 
%In the following, we describe these limitations and propose our extensions to address them.  
%\paragraph{Lexicalizing entity transitions:} 
It does not consider any lexical information regarding the entities, thus, cannot distinguish between transitions of different entities. Although the extended neural grid model proposed in \cite{dat-joty:2017} does incorporate entity features like named entity type and proper mention, it requires an explicit feature extraction step using tools like named entity recognizer. This can prevent us in transferring the model to resource-poor languages or domains.

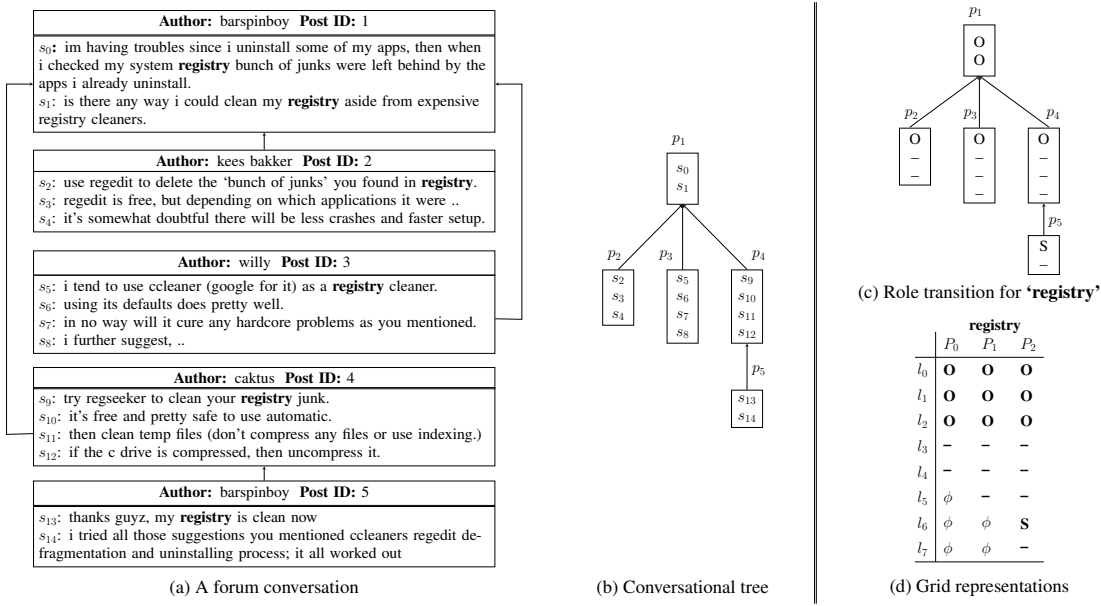
\begin{figure*}
\raggedright
\resizebox{0.97\linewidth}{!}{
\small{
\setlength{\tabcolsep}{4em}
\begin{tabular}{ll||l}
\begin{tikzpicture}[>=stealth, thick]
\LARGE{
\node (A1) at (0,1.6) [draw, process, text width=17.5cm, minimum height=0.5cm, align=center] 
{\textbf{Author:} barspinboy \textbf{ Post ID:} 1};

\node (A2) at (0,-0.8) [draw, process, text width=17.5cm, minimum height=0.7cm, align=left] 
{ \textbf{$s_0$:} im having troubles since i uninstall some of my apps, then when i checked my system \textbf{registry} bunch of junks were left behind by the apps i already uninstall.\\ 
$s_1$: is there any way i could clean my \textbf{registry} aside from expensive registry cleaners. };
%my computer suffers from crash and slow start up. please give me advice so i can clear up ...};

\node (B1) at (0,-3.80) [draw, process, text width=17.5cm, minimum height=0.5cm, align=center] 
{\textbf{Author:} kees bakker \textbf{ Post ID:} 2};

\node (B2) at (0,-5.4) [draw, process, text width=17.5cm, minimum height=0.7cm, align=left] 
{
$s_2$: use regedit to delete the `bunch of junks' you found {in \textbf{registry}}.\\ 
$s_3$: regedit is free, but depending on which applications it were .. \\
$s_4$: it's somewhat doubtful there will be less crashes and faster setup.};

\node (C1) at (0,-7.8) [draw, process, text width=17.5cm, minimum height=0.5cm, align=center] 
{\textbf{Author:} willy \textbf{ Post ID:} 3};

\node (C2) at (0,-9.8) [draw, process, text width=17.5cm, minimum height=0.7cm, align=left] 
{$s_5$: i tend to use ccleaner (google for it) as a \textbf{registry} cleaner. \\ 
$s_6$: using its defaults does pretty well.  \\
$s_7$: in no way will it cure any hardcore problems as you mentioned. \\
$s_8$: i further suggest, .. };

\node (D1) at (0,-12.3) [draw, process, text width=17.5cm, minimum height=0.5cm, align=center] 
{\textbf{Author:} caktus \textbf{ Post ID:} 4};
\node (D2) at (0,-14.2) [draw, process, text width=17.5cm, minimum height=0.7cm, align=left] 
{ $s_{9}$: try regseeker {to clean your \textbf{registry} junk}. \\ 
$s_{10}$: it's free and pretty safe to use automatic.  \\
$s_{11}$: then clean temp files (don't compress any files or use indexing.)  \\
$s_{12}$: if the c drive is compressed, then uncompress it.};

\node (E1) at (0,-16.8) [draw, process, text width=17.5cm, minimum height=0.5cm, align=center] 
{\textbf{Author:} barspinboy \textbf{ Post ID:} 5 };

\node (E2) at (0,-18.5) [draw, process, text width=17.5cm, minimum height=0.7cm, align=left ] 
{$s_{13}$: thanks guyz, {my \textbf{registry} is clean now} \\
$s_{14}$: i tried all those suggestions you mentioned ccleaners regedit defragmentation and uninstalling process; it all worked out};

\coordinate (x) at (-10.0,-14.5);
\coordinate (y) at (-10.0,-1.5);

\coordinate (z) at (10.0,-1.5);
\coordinate (k) at (10.0,-10);

\draw[->] (B1) -- (A2);
\draw[->] (E1) -- (D2);

\draw[->] (C2) -- (k) -- (z) |- (A2);
\draw[->] (D2) -- (x) -- (y) |- (A2);

\node (t) at (0, -20.5) {\huge{(a) A forum conversation}};
}
\end{tikzpicture} 

&
\LARGE{
\begin{tikzpicture}[>=stealth, thick]

\node (P1) at (0,11.5) [text width=0.8cm] {$p_1$};

\node (A1) at (0,10) [draw, process, text width=0.8cm, minimum height=2.0cm, align=center] 
{$s_0$\\
{$s_1$}};

\node (P2) at (-2.5,7) [text width=0.8cm] {$p_2$};

\node (A2) at (-2.5,5.35) [draw, process, text width=0.8cm, minimum height=0.5cm, align=center] 
{$s_2$\\ 
$s_3$\\
$s_4$};

\node (P3) at (-0.5,7) [text width=0.8cm] {$p_3$};
\node (A3) at (0,5) [draw, process, text width=0.8cm, minimum height=0.5cm, align=center] 
{$s_5$\\ 
$s_6$\\
$s_7$\\
$s_8$};

\node (P4) at (3.1,7) [text width=0.8cm] {$p_4$};
\node (A4) at (2.5,5) [draw, process, text width=0.8cm, minimum height=0.5cm, align=center] 
{$s_9$\\ 
$s_{10}$\\
$s_{11}$\\
$s_{12}$};

\node (P4) at (3.1,2.5) [text width=0.8cm] {$p_5$};

\node (A5) at (2.5,1) [draw, process, text width=0.8cm, minimum height=0.5cm, align=center] 
{$s_{13}$\\ 
$s_{14}$};

\coordinate (x) at (0,9);
\coordinate (y) at (-2.5,6.45);
\coordinate (z) at (2.5,6.45);
\draw[<-] (x) -- (y);
\draw[<-] (x) -- (z);

%\coordinate (z) at (4.5,0);
%\coordinate (k) at (4.5,-7.7);

\draw[<-] (A1) -- (A3);
%\draw[-] (A1) -- (A3);
%\draw[-] (A1) -- (A4);
\draw[<-] (A4) -- (A5);

%\draw[-] (A1) -- (x);
%\draw[->] (E1) -- (D2);

%\draw[->] (C2) -- (k) -- (z) |- (A2);
%\draw[->] (D2) -- (x) -- (y) |- (A2);

\node (t) at (0, -6) {\huge{(b) Conversational tree}};

\end{tikzpicture}
}
&
\LARGE{
\begin{tikzpicture}[>=stealth, thick]

\node (P1) at (0,16.5) [text width=0.8cm] {$p_1$};

\node (A1) at (0,15) [draw, process, text width=0.8cm, minimum height=2.0cm, align=center] 
{O\\
{O}};

\node (P2) at (-2.5,12.5) [text width=0.8cm] {$p_2$};

\node (A2) at (-2.5,10.85) [draw, process, text width=0.8cm, minimum height=0.5cm, align=center] 
{O\\ 
--\\
--};

\node (P3) at (-0.2,12.5) [text width=0.8cm] {$p_3$};
\node (A3) at (0,10.5) [draw, process, text width=0.8cm, minimum height=0.5cm, align=center] 
{O\\ 
--\\
--\\
--};

\node (P4) at (3.0,12.5) [text width=0.8cm] {$p_4$};
\node (A4) at (2.5,10.5) [draw, process, text width=0.8cm, minimum height=0.5cm, align=center] 
{O\\ 
--\\
--\\
--};

\node (P4) at (3.1,8.3) [text width=0.8cm] {$p_5$};

\node (A5) at (2.5,7.0) [draw, process, text width=0.8cm, minimum height=0.5cm, align=center] 
{S\\ 
--};

\coordinate (x) at (0,14);
\coordinate (y) at (-2.7,12.0);
\coordinate (z) at (2.7,12.0);
\draw[<-] (x) -- (y);
\draw[<-] (x) -- (z);

%\coordinate (z) at (4.5,0);
%\coordinate (k) at (4.5,-7.7);

\draw[<-] (A1) -- (A3);
%\draw[-] (A1) -- (A3);
%\draw[-] (A1) -- (A4);
\draw[<-] (A4) -- (A5);

%\draw[-] (A1) -- (x);
%\draw[->] (E1) -- (D2);

%\draw[->] (C2) -- (k) -- (z) |- (A2);
%\draw[->] (D2) -- (x) -- (y) |- (A2);

\node (t) at (0, 5.5) {\huge{(c) Role transition for \textbf{`registry'}}};

%Grid Representation

\node (P0) at (0,4.2) [text width=0.8cm] {{\bf registry}};

\node (P1) at (2, 3.5) [text width=0.8cm] {$P_2$};
\node (P1) at (0.5, 3.5) [text width=0.8cm] {$P_1$};
\node (P1) at (-1, 3.5) [text width=0.8cm] {$P_0$};

\node (P1) at (-2, 2.5) [text width=0.8cm] {$l_0$};
\node (P1) at (2, 2.5) [text width=0.8cm] {{\bf O}};
\node (P1) at (.5, 2.5) [text width=0.8cm] {{\bf O}};
\node (P1) at (-1, 2.5) [text width=0.8cm] {{\bf O}};

\node (P1) at (-2, 1.5) [text width=0.8cm] {$l_1$};
\node (P1) at (2, 1.5) [text width=0.8cm] {{\bf O}};
\node (P1) at (.5, 1.5) [text width=0.8cm] {{\bf O}};
\node (P1) at (-1, 1.5) [text width=0.8cm] {{\bf O}};

\node (P1) at (-2, .5) [text width=0.8cm] {$l_2$};
\node (P1) at (2, .5) [text width=0.8cm] {{\bf O}};
\node (P1) at (.5, .5) [text width=0.8cm] {{\bf O}};
\node (P1) at (-1, .5) [text width=0.8cm] {{\bf O}};

\node (P1) at (-2, -.5) [text width=0.8cm] {$l_3$};
\node (P1) at (2, -.5) [text width=0.8cm] {{\bf --}};
\node (P1) at (.5, -.5) [text width=0.8cm] {{\bf --}};
\node (P1) at (-1, -.5) [text width=0.8cm] {{\bf --}};

\node (P1) at (-2, -1.5) [text width=0.8cm] {$l_4$};
\node (P1) at (2, -1.5) [text width=0.8cm] {{\bf --}};
\node (P1) at (.5, -1.5) [text width=0.8cm] {{\bf --}};
\node (P1) at (-1, -1.5) [text width=0.8cm] {{\bf --}};

\node (P1) at (-2, -2.5) [text width=0.8cm] {$l_5$};
\node (P1) at (2, -2.5) [text width=0.8cm] {{\bf --}};
\node (P1) at (.5, -2.5) [text width=0.8cm] {{\bf --}};
\node (P1) at (-1, -2.5) [text width=0.8cm] {$\phi$};

\node (P1) at (-2, -3.5) [text width=0.8cm] {$l_6$};
\node (P1) at (2, -3.5) [text width=0.8cm] {{\bf S}};
\node (P1) at (.5, -3.5) [text width=0.8cm] {$\phi$};
\node (P1) at (-1, -3.5) [text width=0.8cm] {$\phi$};

\node (P1) at (-2, -4.5) [text width=0.8cm] {$l_7$};
\node (P1) at (2, -4.5) [text width=0.8cm] {{\bf --}};
\node (P1) at (.5, -4.5) [text width=0.8cm] {$\phi$};
\node (P1) at (-1, -4.5) [text width=0.8cm] {$\phi$};

\coordinate (x3) at (-1.7, 4);
\coordinate (y3) at (-1.7,-5);
\draw[-] (x3) -- (y3);

\coordinate (x4) at (-2.5,3);
\coordinate (y4) at (2.5, 3);
\draw[-] (x4) -- (y4);

\coordinate (x5) at (-2.5,-5.0);
\coordinate (y5) at (2.5, -5.0);
\draw[-] (x5) -- (y5);

\node (t) at (0, -6) {\huge{(d) Grid representations}};

\end{tikzpicture}
}

\end{tabular}
}
}
\vspace{-0.1em}
\caption{{(a) A forum conversation, (b) Thread structure of the conversation, (c) Entity role transition over a conversation tree}, and (d) 2D role transition matrix for an entity; $\phi$ denotes zero-padding.}
%; $p_k$ denotes post with id $k$, and $s_i$ denotes $i$-th sentence in the temporal order of the conversation.}}
%\vspace{-0.6em}
\label{fig:thread-tree}
\end{figure*}

To address this limitation, we propose to lexicalize entity transitions. This can be achieved by attaching the entity with the grammatical roles. For example, if an entity $e_j$ appears as a subject (S) in sentence $s_i$, the grid entry $E_{i,j}$ will be encoded as \textsc{$e_j$-s}. This way, an entity \textsc{obama} as subject (\textsc{obama-s}) and as object (\textsc{obama-o}) will have separate entries in the embedding matrix $M$. We can initialize the word-role embeddings randomly, or with pre-trained embeddings for the word (\textsc{obama}). In another variation, we kept word and role embeddings separate and concatenated them after the look-up, thus enforcing \textsc{obama-s} and \textsc{obama-o} to share a part of their representations. However, in our experiments, we found the former approach to be more effective.

%(\eg\ word2vec \cite{Mikolov.Sutskever:13}) 
%\red{\paragraph{Applying multiple filters and batch normalization:} The existing model employs only one convolution operation at a time, thus cannot extract and combine features from varying window sizes. In our extended model, we concurrently employ multiple filters of different window sizes. In addition, we apply \emph{batch normalization} to optimize network training \cite{Ioffe:2015:BNA}.}

%\begin{itemize}
%\item Implement the extension with word types (using shared word embeddings). 
%\item Visualize embeddings for shared vs. non-shared
%\item See effect of different types of pooling? 
%\item Re-evaluate the variable margin loss
%\end{itemize}

\section{Coherence Models for Asynchronous Conversations}
\label{sec:coh-model-conv}

%\input{figures/thread-tree-example}

%\vspace{-0.3em}
The main difference between monologue and asynchronous conversation is that information flow in asynchronous conversation is not sequential as in monologue, {rather it is} often interleaved. For example, consider the forum conversation in Figure \ref{fig:thread-tree}(a). There are three possible subconversations, each corresponding to a path {from the root node to a leaf node} in the conversation graph in Figure \ref{fig:thread-tree}(b). In response to seeking suggestions about how to clean \emph{system registry}, the first path ($p_1$$\leftarrow$$p_2$) suggests to use \emph{regedit}, the second path ($p_1$$\leftarrow$$p_3$) suggests \emph{ccleaner}, and the third one ($p_1$$\leftarrow$$p_4$) suggests using \emph{regseeker}. These discussions are interleaved in the chronological order of the posts ($p_1$$\leftarrow$$p_2$$\leftarrow$$p_3$$\leftarrow$$p_4$$\leftarrow$$p_5$). Therefore, monologue-based coherence models may not be effective if applied directly to the conversation.

%If we consider the temporal order of the posts $p_1 \leftarrow p_2 \leftarrow p_3 \leftarrow p_4  \leftarrow p_5$,  As a consequence, coherence models that are originally developed for monologue may not perform as expected when they are directly applied to this conversation. 

We hypothesize that coherence models for asynchronous conversation should incorporate the conversational structure like the tree structure in Figure \ref{fig:thread-tree}(b), where the nodes represent posts and the edges represent `reply-to' links between them. Since the  grid models operate at the sentence level, we construct conversational structure at the sentence level. We do this by linking the boundary sentences across posts and by linking sentences in the same post chronologically. Specifically, we connect the first sentence of post $p_j$ to the last sentence of post $p_i$ if $p_j$  replies to $p_i$, and sentence $s_{t+1}$ is linked to $s_t$ if both $s_{t}$ and $s_{t+1}$ are in the same post.\footnote{The links between sentences are not explicitly shown in Figure \ref{fig:thread-tree}(b) to avoid visual clutter.} Now the question is, 
how can we represent a conversation tree with an entity grid, and then model entity transitions in the tree? In the following, we describe our approach to {this problem}.  

%In our corpus, ``reply-to'' links between comments are given as meta data. However, in many other settings these pointers are unavailable (\eg\ conversations in \href{https://www.quora.com/}{Quora}) or partially available (\eg\  MOOC forums in \href{https://www.coursera.org/}{Coursera}). In such scenarios, we can rely on predictive models to construct the thread structure from the comments of a conversation (more on this later in Section \ref{subsec:thread-recon}). Now the question is how can we represent a conversation tree with an entity grid? In the following, we describe two methods to do this.  

%\vspace{-0.2em}

\subsection{Conversational Entity Grid}
\label{subsec:conv-egrid}

The conversation tree captures how topics flow in an asynchronous conversation. %Each path from the root to a leaf node in the tree can be interpreted as a subconversation covering a subtopic. 
Our key hypothesis is that in a coherent conversation entities exhibit certain local patterns in the conversation tree in terms of their distribution and syntactic realization. Figure \ref{fig:thread-tree}(c) shows how the grammatical roles of entity \emph{`registry'} in our example conversation change over the  tree. For coherence assessment, we wish to model entity transitions along each of the  conversation paths (top-to-bottom), and also their spatial relations across the paths (left-to-right). The existing grid representation is insufficient to model the \emph{two-dimensional (2D) spatial} entity transitions in a conversation tree.  

%%%%%%%%%%%%%%%%%%%%%%%%%%
\begin{figure*}[t]
\centering
\includegraphics[width=0.80\textwidth]{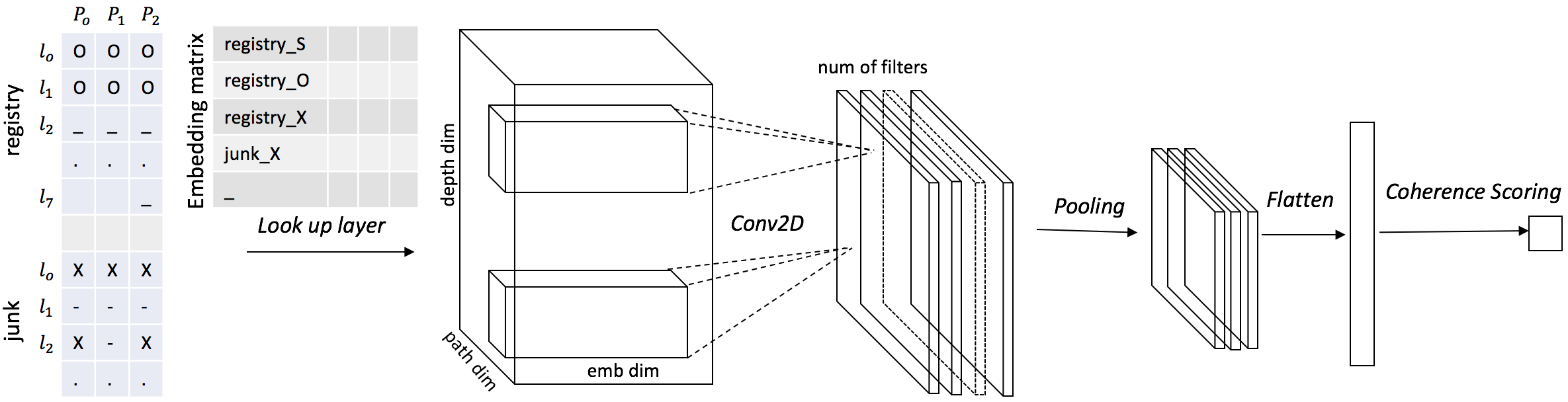}
\vspace{-0.5em}
\caption{\textbf{Conversational Neural Grid} model for assessing coherence in asynchronous conversations.}
%\caption{\red{Dat: You can get rid of line $l_7$ if you want to save some space. Write embedding matrix vertically at the left of the matrix. Put depth dim like before (not inside), may be slightly lower. Put coherence scoring, not just scoring..}}
\label{fig:cnn_model}
%\vspace{-0.1em}
\end{figure*}
%%%%%%%%%%%%%%%%%%%%%%%%%%

%\input{figures/entity-trans-tree-example.tex}

{We propose a three-dimensional (3D) grid for representing entity transitions in an asynchronous conversation. The first dimension  in our grid represents \emph{entities}, while the second and third dimensions represent \emph{depth} and \emph{path} of the tree, respectively. Figure \ref{fig:thread-tree}(d) shows an example representation for {an} entity `\emph{registry}'. Each column in the matrix represents transitions of the entity along a path, whereas each row represents transitions of the entity at a level of the conversation tree.} 

{Although illustrated with a tree structure, our method is applicable to general graph-structured conversations, where a post can reply to multiple previous posts. Our model relies on paths from the root to the leaf nodes, which can be extracted for any graph as long as we avoid loops.}

%\vspace{-0.1em}
\subsection{Modeling Entity Transitions}
\label{subsec:model-tran}

%\vspace{-0.0em}
As shown in Figure \ref{fig:cnn_model}, given a 3D entity grid as input, the look-up layer (Eq. \ref{lookup}) of our neural grid model produces a 4D tensor $L$$\in$$\real^{I \times J \times P \times d}$, where $I$ is the total number of entities in the conversation, $J$ is the depth of the tree, $P$ is the number of paths in the tree,  and $d$ is the embedding dimension. The convolution layer then uses a 2D {filter} $\mathbf{w} \in \real^{m.n.d}$ to convolve local patches of entity transitions  

%\vspace{-1em}
\begin{equation}
%\vspace{-1em}
z_i = h(\mathbf{w}^T L_{i,j:j+m, p:p+n} + {b}_i)
%\vspace{-0.2em}
\end{equation}  

\noindent where $m$ and $n$ are the height and width of the filter, and $L_{i, j:j+m, p:p+n} \in \real^{m.n.d}$ denotes a concatenated vector containing $(m \times n)$ embeddings representing a 2D window of entity transitions. As we repeatedly apply the filter to each possible window with stride size $1$, we get a 2D feature map $Z^i$ of dimensions $(I.J+m-1) \times ( I.P+n-1)$. %\footnote{We use a wide convolution by zero-padding.} 
Employing $N$ different filters, we get $N$ such 2D feature maps, $[{Z}^1, \cdots, {Z}^N]$, based on which the max pooling layer extracts the most salient features:      

%\vspace{-1.0em}
\begin{equation}
\mathbf{p} = [\mu_{l \times w}({Z}^1), \cdots, \mu_{l \times w}({Z}^N)] \label{eq:max_pool_conv}
%\vspace{-0.2em}
\end{equation}

\noindent where $\mu_{l \times w}$ refers to the $\max$ operation applied to each non-overlapping 2D window of $l \times w$ features in a feature map. The pooled features are then linearized and used for coherence scoring in the final layer of the network as described by Equation \ref{dense}.

\section{Experiments on Monologue}
\label{sec:exp}
To validate our proposed extension to the neural grid model, we first evaluate our lexicalized neural grid model in the standard evaluation setting.    

%\vspace{-0.2em}
\paragraph{Evaluation Tasks and Dataset:} \label{subsec:disc-mono}

We evaluate our models on the standard \textbf{discrimination} task \cite{Barzilay:2008}, where a coherence model is asked to distinguish an original document from its incoherent renderings generated by random permutations of its sentences. The model is considered correct if it ranks the original document higher than the permuted one. 

%Following previous work, we use $20$ permutations of each document in the test set. 

%\vspace{-1em}
%\paragraph{Dataset:} 
We use the same train-test split of the \wsj\ dataset as used in \cite{dat-joty:2017} and other studies \cite{Elsner:2011,Feng:2014}. %,,
Following previous studies, we use $20$ random permutations of each article for both training and testing, and exclude permutations that match the original article. Table \ref{table:data} gives some statistics about the dataset along with the number of pairs used for training and testing. \citet{dat-joty:2017} randomly selected $10\%$ of the training pairs for development purposes, which we also use for tuning hyperparameters in our models.

{In addition to the standard setting, we also evaluate our models on an \emph{inverse-order} setting, where we ask the models to distinguish an original document from the inverse order of its sentences (\ie\ from last to first). The  transitions of roles in a negative grid are in the reverse order of the original grid. We do not train our models explicitly on this task, rather use the trained model from the standard setting. The number of test pairs in this setting is same as the number of test documents.} 

%\red{This test validates if a model can distinguish ordering differences } 

%\blue{A second experiment that would make more sense would be to provide as input the sentences in the inverse order (from last to first)}

%\footnote{We received from them in a personal communication.} 

%  
%; see \# Pairs in Table \ref{table:data} for the resulting number of (\emph{original}, \emph{permuted}) pairs.  

%In the {insertion} task \cite{Elsner:2008,Elsner:2011}, the models are judged based on their ability to locate the original position of a sentence in the document from which it was previously removed. To measure this, each sentence in the document is removed and reinserted into every position at a time, and the model is asked to evaluate each such candidate orderings of the document. An insertion place is proposed for which the model gives the highest coherence score to the document. The overall insertion score is then computed as the average fraction of sentences per document reinserted in their original position.

%\vspace{-1em}
\begin{table}[tb!]
%\ra{0.70}
\resizebox{0.94\columnwidth}{!}{%
\begin{tabular}{l|cccc}
\toprule
		& Sections & \# Doc.  & Avg. \# Sen. & \# Pairs  \\
\midrule
{Train} & {00-13}  & 1,378 & 21.5 & 26,422 \\
{Test}  & {14-24}  & 1,053 & 22.3 & 20,411 \\
\bottomrule 
\end{tabular}
}
\vspace{-0.1em}
\caption{Statistics on the \textsc{wsj} dataset.}
\label{table:data}
\vspace{-0.5em}
\end{table}

%\vspace{-0.3em}
\paragraph{Model Settings and Training:}

We train the neural models with the pairwise ranking loss in Equation \ref{loss}. For a fair comparison, we use similar model settings as in \cite{dat-joty:2017}\footnote{\href{https://ntunlpsg.github.io/project/coherence/n-coh-acl17/}{https://ntunlpsg.github.io/project/coherence/n-coh-acl17}} 
      -- ReLU as activation functions ($h$), RMSprop \cite{Tieleman12} as the learning algorithm, Glorot-uniform \cite{GlorotAISTATS2010} for initializing weight matrices, and uniform $\Us(-0.01,  0.01)$ for initializing embeddings randomly. We applied {batch normalization} \cite{Ioffe:2015:BNA}, which gave better results than using dropout. {Minibatch} size, {embedding} size and filter number  were fixed to 32, 300 and 150, respectively. We tuned for optimal filter and pooling lengths in $\{2, \cdots, 12 \}$. We train up to 25 epochs, and select the model that performs best on the development set; see \textbf{supplementary} documents for best hyperparameter settings for different models.
%and %Table \ref{table:ordering} shows the optimal \textbf{win}dow \textbf{size} for our models.  
%obtained best results for $\{9, 10, 11 \}$ on the validation set; 
%\red{see Appendix for our best hyperparameter settings.} %The best model on the development set is then used for the final evaluation on the test set.
We run each experiment five times, each time with a different random seed, and we report the average of the runs to avoid any randomness in results. Statistical significance tests are done using an \emph{approximate randomization} test {with} SIGF V.2 \cite{sigf06}. %with 10,000 iterations. 

\paragraph{Results and Discussions:}

{We present our results on the standard discrimination task and the inverse-order task  in Table \ref{table:ordering}; see Std ($F_1$) and Inv ($F_1$) columns, respectively. For space limitations, we only show $F_1$ scores here, and report both accuracy and $F_1$ in the supplementary document.} We compare our lexicalized models (group III) with the unlexicalized models (group II) of \citet{dat-joty:2017}.\footnote{Our reproduced results for the neural grid model are slightly lower than their reported results ($\sim$ 1\%). We suspect this is due to the randomness in the experimental setup.} We also report the results of non-neural entity grid models \cite{Elsner:2011} in group I. The extended versions use entity-specific features. %The Random baseline at the top makes a random discrimination decision.

\begin{table}[tb!]
\resizebox{0.99\columnwidth}{!}{%
\begin{tabular}{cl|ccc}
\toprule 
& Model & Emb. & \textbf{Std} (${F_1}$)  & {\textbf{Inv} ($F_1$)} \\
\midrule 
%I & Random & - & 50.0 & 50.0 \\
%\midrule 
\multirow{2}{*}{I} 
 & \gridallnoun\ (E\&C) & - & 81.60 & 75.78  \\
 & Ext. Grid (E\&C) & - & 84.95 & 80.34  \\
\midrule
\multirow{2}{*}{II} 
& \gridcnn\ (N\&J) & Random &  84.36 & 83.94 \\
& Ext. \gridcnn\ (N\&J) & Random & {85.93} & 83.00 \\
\midrule 
\multirow{2}{*}{III} 
& \gridcnnlex & Random  & 87.03$^\dagger$ & 86.88$^\dagger$ \\ 
 & \gridcnnlex & Google  &  \textbf{88.56}$^\dagger$ & \textbf{88.23}$^\dagger$\\ 
%\gridcnnlex & Google & \textbf{89.00}  & \textbf{89.00} \\ 
%\red{\gridcnn$_{emb}$} &  & \red{shared Glove}  & 82.29*  & 82.29*  & ? \\ 
\bottomrule 
\end{tabular}
}
\vspace{-0.3em}
\caption{{Dis}crimination results on the \textbf{\wsj} dataset. Superscript $^\dagger$ indicates a lexicalized model is significantly superior to the unlexicalized  \gridcnn\ (N\&J) model with p-value $<0.01$.}
\label{table:ordering}
\vspace{-0.3em}
\end{table}

We experimented with both \emph{random} and \emph{pre-trained} initialization for word embeddings in our lexicalized models. As can be noticed in Table \ref{table:ordering}, both versions give significant improvements over the unlexicalized models on both the standard and the inverse-order discrimination tasks (2.7 - 4.3\% absolute). Our best model with Google pre-trained embeddings \cite{Mikolov.Sutskever:13} yields state-of-the-art results. {We also experimented with Glove  \cite{pennington2014glove}, which has more vocabulary coverage than word2vec -- Glove covers $89.77\%$ of our vocabulary items, whereas word2vec covers $85.66\%$. However, Glove did not perform well giving $F_1$ score of $86\%$ in the standard discrimination task. \citet{schnabel2015} also report similar results where word2vec was found to be superior to Glove in most evaluation tasks.} %After excluding the less frequent (10%) words, our lexical vocabulary contains 17,913 words, of which Glove covers 16,081  and word2vec covers 15,345 (85.66%). Thus, coverage is not an issue. 
%Please see http://blog.aylien.com/overview-word-embeddings-history-word2vec-cbow-glove/ for an informal overview. 
Our model also outperforms the extended neural grid  model that relies on an additional feature extraction step for entity features. These results demonstrate the efficacy of lexicalization in capturing fine-grained entity information without loosing generalizability, thanks to distributed representation and pre-trained embeddings.

\section{Experiments on Conversation}
\label{sec:exp2}
We evaluate our coherence models for asynchronous conversations on two tasks: discrimination and thread reconstruction.
%, as described below. 

%-- discrimination and thread reconstruction. %, as we describe below. 

\subsection{Evaluation on Discrimination}

The discrimination tasks are applicable to conversations also. We first present the dataset we use, then we describe how we create coherent and incoherent examples to train and test our models.

%Similar to \citet{dat-joty:2017}, we train our conversational neural grid model using a pairwise ranking approach. We do not alter the tree structure for negative examples. Both original and permuted conversations share the same tree structure with position of the sentences changed based on the permutation.

%\vspace{-0.3em}
\paragraph{Dataset:}

Our conversational corpus contains discussion threads regarding \emph{computer troubleshooting} from the technology related news site {CNET}.\footnote{https://www.cnet.com/} This corpus was originally collected by \citet{louis2015conversation}, and it contains 13,352 threads. For our experiments, we selected 3,825 threads assuring that each contains at least 3 and at most 15 posts. We use 2,400 threads for training, 750 for testing and 675 for development purposes. Table \ref{table:corpora} shows some basic statistics about the resulting dataset. The threads roughly contain 29 sentences and 6 comments on average. 

%, and about 58\% of these comment reply to comment that is not the first comment in the conversation.      

%\vspace{-0.3em}
\paragraph{Model Settings and Training:}
\label{subsec:train}

To validate the efficacy of our conversational  grid model, we compare it with the following baseline settings:

%\vspace{-0.5em}
\begin{noindlist}\setlength\itemsep{-0.0em}
\item \textbf{Temporal:} In the temporal setting, we construct an entity grid from the chronological order of the sentences in a conversation, and use it with our monologue-based coherence models. Models in this setting thus disregard the structure of the conversation and treat it as a monologue.

\item \textbf{Path-level:} This is a special case of our model, where we consider each path (a column in our conversational grid) in the conversation tree separately. %Recall that a path  (Figure \ref{fig:ent-grid-con}). 
We construct an entity grid for a path and provide as input to our monologue-based models. 
%\red{During testing, we aggregate the path-level win/loss decisions by averaging to estimate the overall coherence of a conversation.}
%\red{During testing, we aggregate the path-level win/loss decisions by averaging to estimate the overall coherence of a conversation.} %Path-level model thus stands in between the temporal model and our conversational grid model in terms of its representation power.
\end{noindlist}
%\vspace{-0.3em}

To train the models with pairwise ranking, we create 20 incoherent conversations for each original conversation by shuffling the sentences in their temporal order. For models involving conversation trees (path-level and our model), the tree structure remains unchanged for original and permuted conversations, only the position of the sentences vary based on the permutation. {Since the shuffling is done globally at the conversation level, this scheme allows us to compare the three representations (temporal, path-level and tree-level) fairly with the same set of permutations.}

\begin{table}[t!]
\resizebox{1.00\columnwidth}{!}{%
\begin{tabular}{l|ccccc}
%\toprule
&  \#Thread & Avg Com & Avg Sen & \#Pairs (tree) & \#Pairs (path)   \\ 
\midrule
{Train} & 2,400  & 6.01  & 28.76 & 47,948  & 106,122\\ 
{Test}  & 750   & 5.75  & 27.79  & 14,986 & 33,852\\ 
{Dev}  & 675    & 6.27  & 30.70  & 13,485 & 28,897\\ 
\midrule
{Total} & 3,825  & 5.98  & 28.77 & 76,419 & 168,871\\ 
\bottomrule
\end{tabular}
}
\vspace{-0.5em}
\caption{Statistics on the \textbf{CNET} dataset.}
\vspace{-0.3em}
\label{table:corpora}
\end{table}

%We exclude permuted examples that match the original in all settings.

{An incoherent conversation may have paths in the tree that match the original paths. We remove those matched paths when training the path-level model. See Table \ref{table:corpora} for number of pairs used for training and testing our models. We evaluate  path-level models by aggregating   correct/wrong decisions for the paths -- if the model makes more correct decisions for the original conversation than the incoherent one, it is counted as a correct decision overall. Aggregating path-level \emph{coherence scores} (\eg\ by averaging or summing) would allow a coherence model to get awarded for assigning higher score to an original path (hence, correct) while making wrong decisions for the rest; see supplementary document for an example. {Similar to the setting in Monologue, we did not train explicitly on the inverse-order task, rather use the trained model from the standard setting.}  

%\footnote{For example, consider a conversation with three paths to which a model assigns 20, 10, and 5 (total 35), and the corresponding incoherent paths get 5, 15, and 10 (total 30). Aggregating scores would favor the model, although it makes wrong decisions for two  out of three.}

%\vspace{-0.3em}
\paragraph{Results and Discussions:}
\label{subsec:discrim-results}

{Table \ref{table:order-results} compares  the results of our models on the two discrimination tasks. We observe more gains in conversation than in monologue for the lexicalized  models -- 4.9\% to 7.3\% on the standard task, and 10\% to 13.6\% on the inverse-order task. Notice especially the huge gains on the inverse-order task. This indicates lexicalization helps to better adapt to new domains.}

{A comparison of the results on the standard task across the representations shows that path-level models perform on par with the temporal models, whereas the tree-level models outperform others by a significant margin. The improvements are 2.7\% for randomly initialized word vectors and 4\% for Google embeddings.} Although, the path-level model considers some  conversational structures, it observes only a portion of the conversation in its input. The common topics (expressed by entities) of a conversation get distributed across multiple conversational paths. This limits the path-level model to learn complex relationships between entities in a  conversation. By encoding an entire conversation into a single grid and by modeling the spatial relations between the entities, our conversational grid model captures both local and global information (topic) of a conversation.

{Interestingly, the improvements are higher on the inverse-order task for both path- and tree-level models. The inverse order yields more dissimilarity at the paths with respect to the original order, thus making them easier to distinguish.}

%\red{Comment on filter width and its implications}

%ranging from 4.9\% to 7.3\% in $F_1$ across the  representations. 
%in all cases conversational structure improves the performance giving $x.x - y.y\%$ absolute gains for path-level and  $x.x - y.y\%$ for our tree-level model. Our tree-level model also outperforms the path-level model by about 2\% in $F_1$. 
%Training the models to discriminate at the path-level may facilitate them with the ability to capture ordering differences at a finer (subtopic) level. 

\begin{table}[tb!]
\resizebox{0.99\columnwidth}{!}{%
\begin{tabular}{llc|cc}
%\toprule
%& & & {\textbf{Dis.}}  \\
Conv. Rep & Model & Emb. & \textbf{Std} ($F_1$) & {\textbf{Inv} ($F_1$)}  \\
%\midrule 
% & Random & - & 50 & 50 \\ 
 \midrule
\multirow{3}{*}{\textbf{Temporal}} 
%& \gridallnoun\ & - & 81.27 & 9.51 \\
%& \gridcnn\ (N\&J)  & - & 81.99\red{?} 82.28 \\
& \gridcnn\ (N\&J)  & random & 82.28 & 70.53\\
& \gridcnnlex & random  & 86.63 & 80.40\\
& \gridcnnlex & Google  & 87.17 & 80.76\\
\midrule
\multirow{3}{*}{\textbf{Path-level}} 
%&\gridcnn\ (N\&J) & random  & 87.86$^\dagger$ \\
&\gridcnn\ (N\&J) & random & 82.39 & 75.68$^\dagger$\\ %84.39 \\
%&\gridcnnlex  & random & 89.46$^\dagger$  \\
&\gridcnnlex  & random & 88.13 & 88.38$^\dagger$ \\
%&\gridcnnlex & Google & 88.93$^\dagger$ \\
&\gridcnnlex & Google & 88.44 & 89.31$^\dagger$ \\
\midrule
\multirow{3}{*}{\textbf{Tree-level}} 
& \gridcnn\ (N\&J)  &  random & 83.98$^\dagger$ & 77.33$^\dagger$ \\ 
%& \gridcnn  & 83.45$^\dagger$ & 83.45$^\dagger$ &  10.56$^\dagger$\\
& \gridcnnlex & random & 89.87$^\dagger$ & 89.23$^\dagger$ \\
& \gridcnnlex & Google & \textbf{91.29}$^\dagger$ & \textbf{90.40}$^\dagger$\\
\bottomrule
\end{tabular}
}
\vspace{-0.2em}
\caption{Discrimination results on \textbf{CNET}. Superscript $\dagger$ indicates a model is significantly superior to its temporal counterpart with p-value $<0.01$.} 
\label{table:order-results}
%\vspace{-0.3em}
\end{table}

\begin{comment}

\begin{table}[tb!]
\resizebox{0.99\columnwidth}{!}{%
\begin{tabular}{llc|cc}
%\toprule
%& & & {\textbf{Dis.}}  \\
Conv. Rep & Model & Emb. & Acc & $F_1$  \\
%\midrule 
% & Random & - & 50 & 50 \\ 
 \midrule
\multirow{3}{*}{\textbf{Temporal}} 
%& \gridallnoun\ & - & 81.27 & 9.51 \\
%& \gridcnn\ (N\&J)  & - & 81.99\red{?} 82.28 \\
& \gridcnn\ (N\&J)  & random & 82.28 & 82.28 \\
& \gridcnnlex & random  & 86.63 & 86.63 \\
& \gridcnnlex & Google  & 87.17 & 87.17 \\
\midrule
\multirow{3}{*}{\textbf{Path-level}} 
%&\gridcnn\ (N\&J) & random  & 87.86$^\dagger$ \\
&\gridcnn\ (N\&J) & random & 81.47  & 82.39 \\ %84.39 \\
%&\gridcnnlex  & random & 89.46$^\dagger$  \\
&\gridcnnlex  & random & 86.13 & 88.13  \\
%&\gridcnnlex & Google & 88.93$^\dagger$ \\
&\gridcnnlex & Google & 86.67 & 88.44 \\
\midrule
\multirow{3}{*}{\textbf{Tree-level}} 
& \gridcnn\ (N\&J)  &  random & 83.98$^\dagger$ & 83.98$^\dagger$ \\ 
%& \gridcnn  & 83.45$^\dagger$ & 83.45$^\dagger$ &  10.56$^\dagger$\\
& \gridcnnlex & random & 89.87$^\dagger$ & 89.87$^\dagger$ \\
& \gridcnnlex & Google & \textbf{91.29}$^\dagger$ & \textbf{91.29}$^\dagger$ \\
\bottomrule
\end{tabular}
}
%\vspace{-0.5em}
\caption{Discrimination results on \textbf{CNET}. Superscript $\dagger$ indicates a model is significantly superior to its temporal counterpart with p-value $<0.01$.} 
\label{table:order-results}
%\vspace{-0.9em}
\end{table}

\end{comment}

If we notice the hyperparameter settings for the best models on this task (see supplementary document), we see they use a filter width of 1. This indicates that to find the right order of the sentences in conversations, it is sufficient to consider entity transitions along the conversational paths in a tree.

%\vspace{-0.3em}
\subsection{Evaluation on Thread Reconstruction}

One crucial advantage of our tree-level model over other models is that we can use it to build predictive models to uncover the thread structure of a conversation from its posts. %, which is not possible with other models. 
 Consider again the thread in Figure \ref{fig:thread-tree}. Our goal is to train a coherence model that can recover the tree structure in Figure \ref{fig:thread-tree}(b) from the sequence of posts $(p_1, p_2, \ldots, p_5)$.  
 
This task has been addressed previously \cite{yi:2008,Wang:2011:PTD}. Most methods learn an edge-level classifier to decide for a possible link between two posts using features like distance in position/time, cosine similarity, etc. To our knowledge, we are the first to use coherence models for this problem. However, our goal in this paper is not to build a state-of-the-art system for thread reconstruction, rather to evaluate coherence models by showing its effectiveness in scoring candidate tree hypotheses. In contrast to previous methods, our approach therefore considers the whole thread structure at once, and computes coherence scores for all possible candidate trees of a conversation. The tree that receives the highest score is predicted as the thread structure of the conversation.

%\vspace{-0.4em}
\paragraph{Training:}

We train our coherence model for thread reconstruction using {pairwise ranking} loss as before. For a given sequence of comments in a thread, we construct a set of valid candidate trees; a valid tree is one that respects the chronological order of the comments, \ie\ a comment can only reply to a comment that precedes it. The training set contains ordered pairs $(T_i, T_j)$, where $T_i$ is a true (gold) tree and $T_j$ is a valid but false tree.  

%We train our best coherence model on conversation from last experiment for the thread reconstruction task.

%Like before, we optimize the following to find model parameters $\theta$ that assign a higher score to $T_i$ than to $T_j$. 
%   This setting allows us to compare our neural grid model with traditional entity grid models on this task.  
        
%\vspace{-0.4em}
\paragraph{Experiments:}
The number of valid trees grows exponentially with the number of posts in a thread, which makes the  inference difficult. 
As a proof of concept that coherence models are useful for finding the right tree, we built a simpler dataset by selecting forum threads from the CNET corpus ensuring that a thread contains at most 5 posts. The final dataset contains 1200 threads with an average of 3.8 posts  and 27.64 sentences per thread. 

%Table \ref{table:recon-corpus} presents the resulting dataset. We use 800 threads for training, 300 for testing, and 100 for development purposes.      
\begin{comment}
\begin{table}[tb]
\centering
\resizebox{0.98\linewidth}{!}{
\begin{tabular}{cccc}
%\toprule
 \# Threads & Avg. \# Posts & Avg. \# Sent & Non-trivial replies  \\ 
\midrule
 1,200  & 3.6 & 27.64 & 57\%\\ 
\bottomrule
\end{tabular}
}
\vspace{-0.5em}
\caption{Forum dataset used for thread reconstruction experiments. \emph{Non-trivial} replies are posts that reply to other posts except the first post.}
\label{table:recon-corpus}
\end{table}
\end{comment}

We assess the performance of the models at two levels: \Ni \textbf{thread-level}, where we evaluate if the model could identify the entire conversation thread correctly, and \Nii \textbf{edge-level}, where we evaluate if the model could identify individual replies correctly. For comparison, we use a number of simple but well performing baselines: 

%\vspace{-0.1em}
\begin{noindlist}\setlength\itemsep{-1.2em}
%\vspace{-0.1em}

\item{\bf All-previous} creates thread structure by linking a comment to its previous (in time) comment.\\

\item{\bf All-first} creates thread structure by linking all the comments to the initial comment.\\

\item{\bf COS-sim} creates thread structure by linking a comment to one of the previous comments with which it has the highest cosine similarity. We use TF.IDF representation for the comments.

%To measure cosine similarity between two comments, each comment is represented with a vector of TF.IDF  scores of its words.  
%(term frequency-inverse document frequency)
%\vspace{-0.5em}
\end{noindlist}

\begin{table}[tb]
\centering
\resizebox{0.95\linewidth}{!}{
\begin{tabular}{l|cll}
%\toprule
 & \textbf{Thread-level} &  \multicolumn{2}{c}{\textbf{Edge-level}} \\
\cmidrule(lr){2-2}\cmidrule(lr){3-4}
 & Acc & $F_1$ & ~Acc \\  
\midrule
%Random    & \\ 
All-previous  & 27.00 & 52.00 & 61.83\\
All-first  & 25.67 & 48.23 & 58.19\\
COS-sim  & 27.66 & 50.56 & 60.30  \\
\midrule
Conv. Entity Grid  & \bf 30.33$^\dagger$ & \bf 53.59$^\dagger$ & \bf 62.81$^\dagger$ \\

%filter-width 3  & 26.33 & 51.15 & 61.44  \\
%filter-width 2  & 28.33 & 53.26 & 63.38  \\
%filter-width 1  & 27.33 & 50.92 & 61.08  \\
  
\bottomrule
\end{tabular}
}
\vspace{-0.2em}
\caption{Thread reconstruction results; $^\dagger$ indicates significant difference from COS-sim (p$<.01$).} 
\label{tab:recon_results}
%\vspace{-0.2em}
\end{table}

Table \ref{tab:recon_results} compares our best conversational grid model (tree-level with Google vectors) with the baselines. The low thread-level accuracy across all the systems prove that reconstructing an entire tree is a difficult task. Models are reasonably accurate at the edge level. Our coherence model shows promising results,  yielding substantial improvements over the baselines. It delivers {2.7\%} improvements in thread-level and {2.5\%} in edge-level accuracy over the best baseline (COS-sim). %These results show promise of coherence models for thread reconstruction task. 

Interestingly, our best model for this task uses a filter width of 2 (maximum can be 4 for 5 posts). This indicates that spatial (left-to-right) relations between entity transitions are important to find the right thread structure of a conversation.

\section{Conclusion}
\label{sec:con}
%\vspace{-0.1em}

We presented a coherence model for asynchronous conversations. We first extended the existing neural grid model by lexicalizing its entity transitions. We then adapt the model to conversational discourse by incorporating the thread structure in its grid representation and feature computation. We designed a 3D grid representation for capturing spatio-temporal entity transitions in a conversation tree, and employed a 2D convolution to compose high-level features from this representation. 

Our lexicalized grid model yields state of the art results on standard coherence assessment tasks in monologue and conversations. We also show a novel application of our model {in} %or with?
forum thread reconstruction. Our future goal is to use the coherence model to generate new conversations.  

%\input{experiments-conv}

%\section*{Acknowledgments}
%The acknowledgments should go immediately before the references.  Do not number the acknowledgments section ({\em i.e.}, use \verb|\section*| instead of \verb|\section|). Do not include this section when submitting your paper for review.

\bibliography{coherence}
\bibliographystyle{acl_natbib}

%\appendix
%\section{Supplemental Material}
%\label{sec:supplemental}

\end{document}